\documentclass[12pt]{article}

\usepackage{amsmath,amsthm}
\usepackage{amsfonts,amssymb}
\usepackage{stmaryrd}

\newtheorem{thm}{Theorem}[section]

\theoremstyle{definition}

\newtheorem{defn}{Definition}[section]

\newcommand  {\bs}		{\backslash}

\renewcommand {\AA}		{\mathsf{A}}

\newcommand	 {\CC}		{\mathsf{C}}

\newcommand  {\OL}		{\overline{\L}}
\newcommand  {\OM}		{\overline{\M}}
\newcommand  {\ON}		{\overline{\N}}
\newcommand  {\OS}		{\overline{\S}}

\newcommand  {\pcat}	{\frown}

\renewcommand{\a}		{\alpha}
\renewcommand{\b}		{\beta}

\newcommand  {\e}		{\epsilon}
\newcommand  {\s}		{\sigma}

\newcommand	 {\A}		{\mathcal{A}}

\renewcommand{\L}		{\mathcal{L}}
\newcommand	 {\M}		{\mathcal{M}}
\newcommand	 {\N}		{\mathcal{N}}

\renewcommand{\P}		{\mathcal{P}}

\renewcommand{\S}		{\mathcal{S}}
\newcommand	 {\T}		{\mathcal{T}}

\newcommand  {\NOT}		{\sim}
\newcommand  {\OR}		{\vee}
\newcommand  {\AND}		{\wedge}

\newcommand  {\imp}		{\rightarrow}

\newcommand  {\SIGN}	{\mathsf{SIGN}}
\newcommand  {\SUBJ}	{\mathsf{SUBJ}}
\newcommand  {\Phon}	{\mathsf{Phon}}
\newcommand  {\Bool}	{\mathsf{Bool}}
\newcommand  {\Prop}	{\mathsf{Prop}}
\newcommand  {\Ind}		{\mathsf{Ind}}
\newcommand  {\true}	{\mathsf{true}}
\newcommand  {\false}	{\mathsf{false}}
\newcommand  {\Sem}	    {\mathsf{Sem}}
\newcommand  {\sem}	    {\mathsf{sem}}
\newcommand  {\phon}	{\mathsf{phon}}

\numberwithin{equation}{section}

\begin{document}     
\vspace{24pt}
\begin{center}
\begin{Large}
\textbf{A Note On Higher Order Grammar\\}
\end{Large}
\vspace{24pt}
Victor Gluzberg\\
HOLTRAN Technology Ltd\\
gluzberg@netvision.net.il
\end{center}

\begin{center}
\textbf{Abstract\\}
\end{center}
Both syntax-phonology and syntax-semantics interfaces in Higher Order Grammar (HOG) are expressed as axiomatic theories in higher-order logic (HOL), i.e. a language is defined entirely in terms of provability in the single logical system. An important implication of this elegant architecture is that the meaning of a valid expression turns out to be represented not by a single, nor even by a few "discrete" terms (in case of ambiguity), but by a "continuous" set of logically equivalent terms. The note is devoted to precise formulation and proof of this observation.

\section{Introduction}

Higher Order Grammar (HOG) \cite{pollard:hana:2003,pollard:2004:cg,pollard:2006} is probably the most recent implementation of the idea of using a single logical system for linguistic generalizations pioneered by \cite{kasper:rounds:1986,king_89,richter:diss} and, at the same time, the first one based on the mainstream classical higher-order logic (HOL), traditionally applied only to the semantics of natural languages \cite{gallin:1975}. Both syntax-phonology and syntax-semantics interfaces in HOG are expressed as axiomatic theories in the HOL, i.e. a language is defined entirely in terms of provability in the single logical system. 

This elegant architecture has an important and almost obvious implication which, however, does not seem to be explicitly mentioned in the literature so far: the meaning of a valid expression turns out to be represented not by a single, nor even by a few "discrete" terms (in case of ambiguity), but by a "continuous" set of logically equivalent terms (though all having a single or a few distinct interpretations in a model of the HOL). The present note is devoted to precise formulation and proof of this observation.

Though HOG is declared to be agnostic about axiomatization of the underlying HOL \cite{pollard:2006}, for our purpose we will assume availability of the description operator, either introduced explicitly as a logical constant, as in $Q_0$ theory of \cite{andrews:1986}, or implied by the description axiom, as in the theory denoted as $Ty_2+D$ by \cite{gallin:1975}. This assumption allows to introduce for every type $\a$ an "if-then-else" constant $\CC_\a:\a \AND \a \AND \Bool \imp \a$ with the following fundamental properties:
\[
\vdash \CC_\a (x_\a, y_\a, \true) = x_a, \qquad
\vdash \CC_\a (x_\a, y_\a, \false) = y_a.
\]
Here and everywhere below we follow notations of \cite{pollard:2006}, with only the two differences: we employ small Greek letters for type variables and $\imp$ as the functional type constructor. We will make use of the following metatheorems
\begin{equation} \label{eq1.1}
\vdash \forall _{f:\a\imp\b} \; f(\CC_\a (x, y, z)) = \CC_\b (f(x), f(y), z)
\end{equation}
\begin{equation} \label{eq1.2}
\vdash x \OR y = \CC_{\Bool} (x, y, x)
\end{equation}
\begin{equation} \label{eq1.3}
\vdash \CC_\a (x, x, z) = x
\end{equation}
which are easily verified under either of $Ty_2+D$ or $Q_0$ axiomatizations.

In the next section we first formalize the notion of a language with HOL-based semantics as an arbitrary relation between the language expressions and HOL terms of a certain type $\a$, referenced as $\a$\emph{-language}. Among examples illustrating this formalization we describe how a HOG defines $\a$-languages for some types $\a$. Then we define an important subclass of $\a$-languages, referenced as \emph{logically closed} languages, and bring some trivial examples of logically closed as well as of non-logically-closed languages. In the section 3 we formally prove any $\a$-language defined by a HOG to be logically closed.

\section{Logically closed languages}

\begin{defn} \label{defn2.1}
Let $\A$ be a finite alphabet $\A = \{ a_1,\:a_2,\:...\:a_N \}$, let $\A^*$ denote the set of all finite words over alphabet $\A$ and let $\T_\a$ denote the set of all the HOL terms of an arbitrary type $\a$. An $\a$-\textit{language} is a relation $\L \subset \A^* \otimes \T_\a$.
\end{defn}

Referring to words over alphabet $\A$ as ``expressions'' and $\a$-terms as ``$\a$-meanings,'' one can say that an $\a$-language is a set of pairs of expressions and their $\a$-meanings. 

\medskip \noindent
\textbf{Examples}

\begin{enumerate}
\item An arbitrary set of words $\L \subset \A^*$ can be considered as a language for the unit type meaning.

\item A trivial particular case of an $\a$-language is a singleton $\{ (w, a:\a) \}$, where $w \in \A^*$.

\item If $\A$ contains all symbols of the HOL own language, so that $\T_\a$ can be identified with a subset of $\A^*$, then the identity relation on $\T_\a$ is an $\a$-language being a subset of the HOL language.

\item A Higher Order Grammar \cite{pollard:2006} with a set ${\{\AA_0,\AA_1,...\:\AA_N\: : \:\Phon\}}$ of phonological constants and a set $\Gamma$ of non-logical axioms about semantics and phonology of specific words and rules of their composition for non-primitive syntactic signs, for every type $\a = \Sem(\s)$, where $\s \in \SIGN$, defines an $\a$-language with an alphabet $\A = \{ a_1,\:a_2,\:...\:a_N \}$ as follows: a pair $(w, a)$ belongs to the language if and only if there exists a sign $s:\s$ such that
\[
\Gamma \vdash \phon(s) = /w/ \:\AND\: \sem(s) = a,
\]
where the mapping $/\cdot/$ is defined by:
\[
/\e/ = \AA_0, \qquad /a_i/ = \AA_i,\; i \in \{1, 2, ... N\}, \qquad /u v/ =  /u/ \pcat /v/
\]
and $u v$ denotes concatenation of words $u$ and $v$. (In fact, \cite{pollard:2006} implicitly applies such a mapping to introduce convenient notations for phonological constants and their concatenations, like 
\[
/\mathsf{fajdo\:blt}/ =_{def} /\mathsf{fajdo}/ \pcat /\mathsf{blt}/).
\]
\end{enumerate}

\begin{defn} \label{defn2.2}
A \textit{logically closed} $\a$-language is an $\a$-language $\L$ such that whenever $(w, b) \in \L, \; (w, c) \in \L$ and $\vdash a = b \OR a = c
$ then $(w, a) \in \L$ also. A minimal logically closed $\a$-language $\OL$ which includes a given arbitrary $\a$-language $\L$ is said to be its \textit{logical closure}.
\end{defn}

This definition actually captures the two important features of an $\a$-language:
\begin{enumerate}
\item If an expression $w$ in the language has a meaning $b$, then it also has every meaning $a$ logically equivalent to $b$
\item If an expression $w$ is ambiguous, i.e. has at least two distinct meanings $b$ and $c$ being \emph{not} logically equivalent, then it also has every meaning $a$ which is provable to be equal either $b$ or $c$.
\end{enumerate}

Thus, every valid expression of a logically closed language is associated not with a single, nor even with a few "discrete" terms (in case of ambiguity), but with a "continuous" set of logically equivalent terms. A precise formulation of this interpretation follows.

\begin{defn} \label{defn2.3}
A set $\M \subset \T_\a$ is said to be \textit{logically closed} if and only if whenever $b \in \M,\; c \in \M$ and
\[
\vdash a = b \OR a = c
\]
then $a \in \M$ also.
A minimal logically closed set $\OM \subset \T_\a$ which includes an arbitrary set $\M \subset \T_\a$ is said to be its \textit{logical closure}.
If in addition, $\N \subset \T_\a$ and $\OM = \ON$, we say the two sets $\M$ and $\N$ are \textit{logically equivalent} and denote this relation as $\M \simeq \N$.
\end{defn}

It is readily seen that $\simeq$ is an equivalence relation in the power set \mbox{$\P(\T_\a)$} and therefore a logically closed $\a$-language might be defined equivalently as a function $\L : \A^* \rightarrow \P(\T_\a)/\simeq$.

The simplest non-empty logically closed $\a$-language is a \textit{logical singleton}
\[
\OS = \{ w \} \otimes \overline{\{ a \}}.
\]

Note that the $\a$-sub-language of the HOL own language is \emph{not}, of course, logically closed.

\section{HOG defined languages are logically closed}

We are now going to focus on the example 4 to Definition \ref{defn2.1} and prove the main result of this note:

\begin{thm} \label{thm3.1}
Any language defined by a HOG is logically closed.
\end{thm}
\begin{proof}
Let $s_1$ and $s_2$ denote two signs of the same type $\s$ (may be, but not necessarily, distinct) which have the same phonology:
\[
\Gamma \vdash \sem(s_1) = a_1:\a,
\]
\[
\Gamma \vdash \sem(s_2) = a_2:\a
\]
\begin{equation} \label{eq3.3}
\Gamma \vdash \phon(s_1) = w,
\end{equation}
\begin{equation} \label{eq3.4}
\Gamma \vdash \phon(s_2) = w,
\end{equation}
and assume a term $a:\a$ be such that
\begin{equation} \label{eq3.5}
\vdash a = a_1 \OR a = a_2
\end{equation}
Consider a sign
\[
s =_{def} \CC_\s (s_1,  s_2, a = a_1).
\]
From \ref{eq3.3}, \ref{eq3.4} by metatheorems \ref{eq1.1} and \ref{eq1.3} it follows that
\[
\Gamma \vdash \phon(s) = w.
\]
Now, by metatheorem \ref{eq1.1}
\begin{equation} \label{eq3.6}
\Gamma \vdash \sem(s) = \CC_\a (a_1, a_2, a = a_1)
\end{equation}
and by metatheorem \ref{eq1.2} the assumption \ref{eq3.5} may be re-written as
\[
\vdash \CC_{\Bool} (a = a_1, a = a_2, a = a_1)
\]
or, again applying \ref{eq1.1},
\[
\vdash \CC_\a (a1, a_2, a = a_1) = a.
\]
Replacing the left-hand term of this equality by its right-hand term in \ref{eq3.6}, we finally obtain
\[
\Gamma \vdash \sem(s) = a.
\]
Thus, the pair $(w, a)$ belongs to the language along with $(w, a_1)$ and $(w, a_2)$.
\end{proof}

\section{Discussion}

We did not consider here the inverse question, i.e. whether any logically closed $\a$-language (of an appropriate type) can be defined by a HOG, basically because its precise statement and resolution may vary depending on whether and how to restrict the set of HOG non-logical constants and primitive $\SIGN$ types and type constructors. This becomes obvious from considering some particular cases like the following. Let a $\Prop$-language contain everything defined by a HOG with the set of primitive $\SIGN$ types and type constructors as in \cite{pollard:2006}, that enforces, in particular, the rule
\[
\forall_{x,f}[\sem(^{\SUBJ} \; x \; f) = \sem(f)(\sem(x))],
\]
except all those pairs $(w, a:\Prop)$ for which there exist signs $s:NP$ and $q:NP\bs_{\SUBJ}S$ such that
\[
\phon(s) \pcat \phon(q) = /w/ \qquad \text{and} \qquad \vdash \NOT p(\sem(s), \sem(q)),
\]
where $p:\Ind\AND(\Ind\imp\Prop)\imp\Bool$ is a given predicate on subject/verb-phrase pairs. It is clear that such a language, being logically closed, in general case can not be defined by a HOG with the same set of primitive $\SIGN$ types and type constructors, just because of the above rule. It might, however, be defined by a HOG with an extended set of primitive types and type constructors.

We also refrain from judgment of up to what extent the limitation of HOG-definable languages by the class of logically closed languages is restrictive or, the opposite, desirable. The goal of this note was only to point out the limitation, as awareness of it is important in any case.

\end{document}